\documentclass[twocolumn,letterpaper]{IEEEAerospaceCLS}

\usepackage[]{graphicx}
\usepackage[mode=match]{siunitx}
\usepackage{amsmath}
\usepackage{booktabs}
\usepackage{algorithm}
\usepackage{algpseudocode}
\usepackage[numbers]{natbib}
\usepackage[colorlinks=true, urlcolor=blue, linkcolor=black, citecolor=black]{hyperref}
\usepackage{cleveref}
\usepackage[caption=false]{subfig}
\newcommand{\ignore}[1]{}

\begin{document}
\title{Scalable LEO Conjunction Screening using Adaptive Synthetic-Covariance Thresholds}
\author{Vedant Srinivas, Grace Ra Kim, Duncan Eddy, and Mykel J. Kochenderfer\\ 
Stanford University\\
496 Lomita Mall \\
Stanford, CA 94305 \\
\{vedants8, gkim65, deddy, mykel\}@stanford.edu
\thanks{\footnotesize
This work has been submitted to the IEEE for possible publication.
Copyright may be transferred without notice, after which this version
may no longer be accessible.}
}
\maketitle
\nocite{IEEEexample:BSTcontrol}

\thispagestyle{plain}
\pagestyle{plain}

\begin{abstract}

\textbf{ As orbital populations grow, conjunction screening must search an increasing number of object pairs to identify those that may pass close enough to pose a collision risk. Existing pipelines rely on inexpensive early-stage filters, such as an apsis filter, to eliminate pairs whose orbital altitude ranges cannot plausibly result in a close approach. However, these filters typically apply the same fixed threshold to every object, making the screening unnecessarily conservative for well-constrained objects while potentially insufficient for objects with greater orbit-prediction uncertainty. In this work, we present an uncertainty-aware conjunction screening method that replaces a single fixed apsis threshold with object-specific thresholds derived from propagated radial uncertainty. The resulting thresholds vary across both objects and prediction time, becoming narrower when propagated uncertainty is small and wider when uncertainty grows. We pair this adaptive screening rule with a binary-search altitude sweep that makes the resulting heterogeneous intervals efficient to evaluate at catalog scale with log-linear search complexity. We evaluate the method on 33,791 objects from the Space-Track General Perturbations (GP) catalog using the preceding eight public ephemeris snapshots and a seven-day prediction horizon. The adaptive sieve retains 162.8 million candidate pairs, reducing the candidate set by 28.78\% relative to a fixed 50 km apsis threshold and by 71.48\% relative to brute-force all-pairs screening. It retains all close approaches identified in two independently propagated reference sets. At the largest tested catalog size, the binary-search sweep is 27 times faster than exhaustive interval evaluation while returning the same candidate set. These results demonstrate that uncertainty-aware screening can substantially reduce the candidate burden passed to downstream conjunction analysis, providing a scalable path toward managing increasingly large and heterogeneous orbital populations.}
\end{abstract}

\tableofcontents

\section{Introduction}

Over 40,000 Earth-orbiting objects are currently tracked by space surveillance networks, and this number is projected to grow substantially to over  100,000 satellites in 2030~\cite{ESA100K2030}. As the orbital population grows, so does the number of potential close approaches, or conjunctions, that could occur between space objects that must be assessed for collision risk. The scale of this problem is already substantial; \citeauthor{Spencer2024} report more than 130,000 conjunctions with separation distances below 5 km in a single week and note that this number is likely to increase with the growing population of operational satellites~\cite{Spencer2024}. NASA has observed the same trend operationally. The number of conjunction events processed by NASA's Conjunction Assessment Risk Analysis (CARA) program remained relatively stable from 2016 through 2020, before increasing sharply as additional payloads entered the orbital regimes of NASA assets~\cite{NASA_CARA_EVENTS}. \citeauthor{motheram_2026} simulated LEO populations increasing from 10,885 to 94,581 satellites and found that the number of conjunctions over a three-month period increased from 8,377 to 511,177 as the population size grew, even without additional debris generation~\cite{motheram_2026}. As orbital populations and conjunction events continue to grow, conjunction-assessment systems must efficiently identify object pairs that warrant further analysis.

Conjunction screening identifies potential close approaches between space objects over a future time window. Given a catalog of objects with their associated orbital elements, screening systems propagate object trajectories and test pairs of objects for separations below a predefined, static miss-distance threshold. In the United States, the U.S. Space Force maintains the Satellite Catalog and distributes orbital data for tracked objects through Space-Track~\cite{SpaceTrack}. These catalog data support conjunction-screening systems such as SOCRATES, which uses publicly available orbital elements and SGP4 propagation to identify potential close approaches~\cite{SOCRATES,vallado_2006}. Conjunction screening enables satellite operators to identify encounters that require further collision-risk assessment, allowing operators to determine when collision-avoidance maneuvers are necessary. The primary challenge in catalog-scale conjunction screening is computational scalability. For a catalog containing $N$ objects, an all-vs-all search contains $N(N-1)/2$ unique pairs, so the number of pairwise comparisons grows as $O(N^2)$. The projected growth of spacecraft populations means that even basic conjunction screening may become computationally impractical, threatening the scalability of downstream collision-risk assessment.

\begin{figure*}[t]
    \centering
    \includegraphics[width=\textwidth]{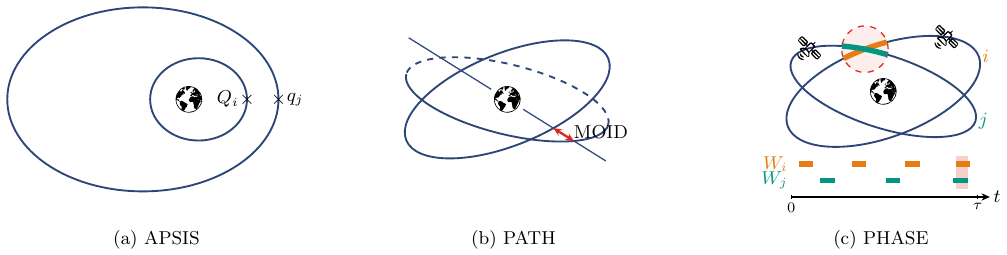}
    \caption{Conjunction-screening filters: (a) apsis, (b) path, and (c) phase.}
    \label{fig:sieve_filters}
\end{figure*}

A common way to reduce the cost of all-vs-all screening is to apply a sequence of conservative filters, or sieves, that remove object pairs from consideration before performing more computationally expensive analysis. \citeauthor{hoots_1984} showed that analytic prefilters could accelerate close-approach detection by eliminating geometrically incompatible satellite pairs~\cite{hoots_1984}. Later work organized these tests into increasingly selective filter sequences that progressively reduce the candidate set~\cite{woodburn_2009,Casanova2014}. The apsis, or altitude, filter is typically applied first and removes pairs whose orbital altitude ranges cannot overlap. Path filters then test whether the two orbital trajectories can approach closely enough, while time or phase filters determine whether the objects can occupy the relevant region of their orbits at approximately the same time, as illustrated in \Cref{fig:sieve_filters}.

These filters are intentionally conservative: they may retain pairs that do not ultimately conjunct, but they aim to avoid discarding pairs that could produce a conjunction. The surviving candidates then proceed to more computationally expensive processing, including higher-fidelity propagation, time-of-closest-approach refinement, and collision-risk analysis. Because the apsis filter is applied first, every pair it removes avoids computation in all subsequent stages, making its filtering efficiency especially impactful at reducing computational cost at catalog scale.

Beyond sieve-based filtering, prior work has addressed conjunction-analysis scalability through improved search algorithms, data structures, machine learning, and parallel computation. \citeauthor{Mercurio2012} use a hierarchical $k$-d tree to reduce the number of object interactions that must be evaluated, while \citeauthor{BudiantoHo2014} use spatiotemporal indexing to restrict conjunction searches to relevant regions of space and time~\cite{Mercurio2012,BudiantoHo2014}. \citeauthor{Hellwig2023} develop spatial data structures designed for parallel CPU and GPU execution~\cite{Hellwig2023}. \citeauthor{stevenson_2023} formulate all-vs-all conjunction screening as a classification problem and evaluate learning-based methods on large sets of candidate pairs~\cite{stevenson_2023}. \citeauthor{motheram_2026} use GPU-parallel propagation and conjunction analysis in the VESTA framework to study simulated LEO populations approaching 100,000 satellites~\cite{motheram_2026}. At a broader scale, MIT's MOCAT framework models the long-term evolution and carrying capacity of the LEO environment, but does not perform object-pair conjunction screening or assess the risk of specific conjunction events~\cite{MOCAT}. These approaches complement conservative geometric filtering at the beginning of the conjunction-analysis pipeline, where reducing the candidate set lowers the computation required by every downstream stage.

The effectiveness of the apsis filter depends on the threshold used to expand each object's radial bounds. Larger thresholds make the filter more conservative but allow more candidate pairs to survive for downstream processing; smaller thresholds improve pair reduction at the cost of greater sensitivity to orbit-prediction and filtering errors~\cite{finkleman_alfano_2014}. Fixed, hand-tuned thresholds are commonly used in practice, with prior works applying a fixed 50~km apsis threshold as the first-stage filter in a catalog-scale all-vs-all conjunction-screening study~\cite{stevenson_2023}. A fixed threshold applies the same padding regardless of the uncertainty of the underlying orbit estimate. For a well-characterized object, the threshold may be unnecessarily large and retain pairs that cannot realistically conjunct. For an object with larger prediction uncertainty, the same threshold may be too small to remain conservative. This mismatch motivates object-specific apsis thresholds that adapt to orbit-prediction uncertainty.

To address this limitation, we introduce an uncertainty-aware apsis filter that replaces the fixed catalog-wide threshold with an object-specific, time-dependent threshold derived from each object's propagated synthetic covariance. We estimate the synthetic covariance by propagating several recent GP element sets for the same object to common epochs, comparing the resulting predicted states with the state predicted from the most recent element set, and using the dispersion of these differences to construct an object-specific covariance estimate. We then propagate this synthetic covariance estimate over the screening horizon to capture how that uncertainty evolves with time. The resulting radial uncertainty determines the apsis threshold for each object. Well-characterized objects receive tighter thresholds, allowing more infeasible pairs to be removed before downstream analysis, while objects with larger prediction uncertainty receive wider thresholds to preserve conservative screening. These thresholds are not capped at 50~km and can widen beyond the fixed baseline when the predicted uncertainty warrants it. The filter therefore allocates screening conservatism according to object-specific uncertainty rather than uniformly across the catalog. 

At catalog scale, evaluating apsis overlap by comparing every pair of altitude intervals becomes increasingly expensive. To make these comparisons tractable, we use a binary-search-based altitude sweep that applies to both fixed and adaptive apsis thresholds. By sorting the intervals and using binary search to identify possible overlaps, the sweep replaces $O(N^2)$ pairwise interval testing with an $O(N\log N)$ search while preserving the same apsis-screening condition.

We evaluate the proposed approach on a snapshot of the Space-Track catalog containing 33,791 valid objects, using the preceding eight public ephemeris snapshots to estimate synthetic covariances where sufficient data are available. At a seven-day look-ahead horizon, the adaptive sieve reduces the 570.9 million possible catalog pairs to 162.8 million candidates, compared with 228.6 million under a fixed 50~km threshold. The adaptive sieve therefore retains 28.78\% fewer candidates than fixed screening and 71.48\% fewer than brute force. Additional experiments evaluate close-approach recall and the computational scaling of the altitude sweep.

\section{Methodology}
\label{sec:methodology}
The proposed method consists of three stages: estimating object-specific prediction uncertainty from past ephemeris data, propagating that uncertainty and using it to construct adaptive apsis filtering thresholds, and evaluating the resulting radial threshold intervals with a binary-search algorithm.

\subsection{Apsis Screening Formulation}

Consider a catalog of \(N\) objects screened over a look-ahead horizon \(\tau\). For each object \(i\), let \(q_i(\tau)\) and \(Q_i(\tau)\) denote the minimum perigee and maximum apogee altitude attained over that horizon. These bounds define the nominal radial envelope
\begin{equation}
A_i(\tau)=[q_i(\tau),Q_i(\tau)]
\label{eq:nominal_radial_envelope}
\end{equation}
To account for orbit-prediction uncertainty, assign object \(i\) a nonnegative radial screening threshold \(T_i(\tau)\). The corresponding expanded interval is
\begin{equation}
I_i(\tau)=
[q_i(\tau)-T_i(\tau),\,Q_i(\tau)+T_i(\tau)]
\label{eq:expanded_radial_interval}
\end{equation}
Objects \(i\) and \(j\) survive the apsis sieve if and only if their expanded radial intervals overlap
\begin{equation}
I_i(\tau)\cap I_j(\tau)\neq\emptyset
\label{eq:apsis_overlap}
\end{equation}
Because the threshold is assigned independently to each object, two otherwise disjoint nominal envelopes can still overlap when their separation is no greater than \(T_i(\tau)+T_j(\tau)\).

The apsis sieve returns a set of candidate pairs that might contain a potential conjunction. However, the radial overlap check applied by the apsis sieve is a necessary condition for a conjunction, but not sufficient on its own\textemdash it does not imply that two objects coincide spatially or temporally over the screening horizon. Therefore, additional filters can be employed to further reduce the candidate set before performing detailed probability of collision or miss-distance computation.

\subsection{Synthetic Covariance Estimation from Public Ephemeris}

Our adaptive threshold requires an estimate of each object's orbit-prediction uncertainty over the conjunction-screening horizon. Ideally, this uncertainty would be obtained from calibrated covariance information produced by an orbit-determination process. Such covariance data, however, are not provided for most objects in the public Space-Track catalog~\cite{vallado_2006,SpaceTrack}. We therefore estimate object-specific synthetic covariances from histories of publicly available General Perturbations (GP) element sets.

Historical GP solutions contain information about prediction disagreement: multiple element sets for the same object can be propagated to common epochs, and differences among the resulting predicted states provide empirical samples from which a covariance-like uncertainty estimate can be constructed. This general approach has been explored in prior work using publicly available SGP4-compatible element sets. \citeauthor{peterson_2001} demonstrated that covariance-like information could be estimated from histories of public element sets~\cite{peterson_2001}; \citeauthor{geul_2017} developed robust weighted differencing methods for characterizing TLE uncertainty and its directional structure in the radial--transverse--normal frame~\cite{geul_2017}; and \citeauthor{oltrogge_2024} developed a digital synthetic-covariance approach based on overlap comparisons between successive orbital solutions~\cite{oltrogge_2024}. Our implementation follows the same general principle of comparing propagated historical solutions, but constructs an object-specific six-dimensional covariance that is subsequently propagated over the screening horizon.

For each object, the ephemeris history is ordered by epoch, with the newest ephemeris defining the reference solution at time \(t_0\). We retain the reference ephemeris and up to \(n_{\mathrm{recent}}=8\) preceding ephemeris states. Each retained ephemeris is propagated with SGP4 over the common prediction interval \([t_0,t_f]\), where \(t_f\) denotes the end of the prediction horizon. We set \(t_f-t_0=7\) days. The sensitivity of the synthetic covariance estimate to the amount of ephemeris history used is evaluated in Appendix~\ref{app:covariance_validation}. At a common sample time \(t\in[t_0,t_f]\), the inertial state difference is
\begin{equation}
\delta\mathbf{x}(t)=
\begin{bmatrix}
\mathbf{r}_{\mathrm{old}}(t)-\mathbf{r}_{\mathrm{ref}}(t)\\
\mathbf{v}_{\mathrm{old}}(t)-\mathbf{v}_{\mathrm{ref}}(t)
\end{bmatrix}
\label{eq:state_residual}
\end{equation}
where \((\mathbf{r}_{\mathrm{old}},\mathbf{v}_{\mathrm{old}})\) is the propagated inertial state from a preceding ephemeris and \((\mathbf{r}_{\mathrm{ref}},\mathbf{v}_{\mathrm{ref}})\) is the propagated inertial state from the reference ephemeris.

Prediction disagreement is direction-dependent because radial error changes the object's orbital radius, transverse error shifts its position along the orbit, and normal error displaces it perpendicular to the orbital plane. We therefore express each residual in the instantaneous radial--transverse--normal (RTN) frame of the reference trajectory. Let \(\mathbf{R}(t)\) denote the inertial-to-RTN rotation matrix at time \(t\). The six-dimensional state prediction residual is then
\begin{equation}
\delta\mathbf{x}^{\mathrm{RTN}}(t)=
\begin{bmatrix}
\mathbf{R}(t)&\mathbf{0}\\
\mathbf{0}&\mathbf{R}(t)
\end{bmatrix}
\delta\mathbf{x}(t)
\label{eq:rtn_residual}
\end{equation}
where the radial component measures disagreement in orbital radius, the transverse component measures disagreement along the primary direction of motion, and the normal component measures disagreement perpendicular to the orbital plane. Although the apsis sieve ultimately uses radial uncertainty, all position and velocity components are retained to preserve the cross-correlations required for covariance propagation.

To characterize uncertainty near the reference epoch, we retain \(M\) finite residual samples uniformly at one-minute intervals over \([t_0,t_0+0.3~\mathrm{days})\), using up to eight preceding ephemeris samples. For \(M\) retained residuals with sample mean \(\bar{\mathbf{x}}^{\mathrm{RTN}}\), the initial synthetic covariance is
\begin{equation}
P_{i,0}^{\mathrm{RTN}}=
\frac{1}{M-1}\sum_{p=1}^{M}
\left(\delta\mathbf{x}_p^{\mathrm{RTN}}-\bar{\mathbf{x}}^{\mathrm{RTN}}\right)
\left(\delta\mathbf{x}_p^{\mathrm{RTN}}-\bar{\mathbf{x}}^{\mathrm{RTN}}\right)^T
\label{eq:initial_covariance}
\end{equation}
where we require at least seven finite residual samples because a mean-centered six-dimensional sample covariance can be full rank only when \(M\geq7\).

The resulting covariance measures local disagreement among recent public orbit estimates rather than operator-grade orbit uncertainty. Residuals generated from the same ephemeris pair are temporally correlated, and both trajectories are propagated using the same SGP4 model, so common modeling bias can cancel during differencing. We therefore interpret \(P_{i,0}^{\mathrm{RTN}}\) as a covariance-like measure of disagreement among public orbit estimates that can be propagated to construct the adaptive screening threshold.

\subsection{Covariance-Based Adaptive Thresholding}

To generate our uncertainty-based screening threshold, we propagate the full six-dimensional covariance numerically rather than fitting a scalar uncertainty-growth law. This preserves the coupling among position and velocity uncertainties and allows the radial uncertainty used by the apsis threshold to evolve with the orbital dynamics. The reference ephemeris state is converted to inertial position and velocity, and the initial covariance \(P_{i,0}^{\mathrm{RTN}}\) is rotated into Earth-centered inertial (ECI) coordinates. The state transition matrix (STM) is integrated with the reference trajectory according to
\begin{equation}
\dot{\boldsymbol{\Phi}}_i(t)=
\mathbf{A}_i(t)\boldsymbol{\Phi}_i(t),
\qquad
\boldsymbol{\Phi}_i(0)=\mathbf{I}
\label{eq:stm_dynamics}
\end{equation}
where \(\mathbf{A}_i(t)\) is the Jacobian of a numerical force model including degree-and-order-6 spherical-harmonic gravity and atmospheric drag, chosen to capture the dominant perturbations affecting LEO trajectories over the screening horizon. The propagated covariance is
\begin{equation}
P_i(t)=
\boldsymbol{\Phi}_i(t,0)
P_{i,0}
\boldsymbol{\Phi}_i(t,0)^T
\label{eq:stm_cov}
\end{equation}
where \(P_i(t)\) is the propagated covariance at time \(t\). We use the \texttt{astrojax} astrodynamics library for GPU-accelerated state and STM propagation~\cite{astrojax}. The numerical propagator uses degree-and-order-6 EGM2008 gravity and a cannonball drag model with the modified Harris--Priester atmospheric density model. For each object, we calibrate an effective drag area to the semimajor-axis decay inferred from its public mean-motion derivative. Given a one-day reference propagation with area \(A_{\mathrm{ref}}=1~\mathrm{m}^2\), the effective area is
\begin{equation}
A_{\mathrm{eff}}
=
A_{\mathrm{ref}}
\max\left(
0,
\frac{(da/dt)_{\mathrm{target}}}
     {(da/dt)_{\mathrm{ref}}}
\right)
\label{eq:effective_drag_area}
\end{equation}
where \((da/dt)_{\mathrm{target}}\) is derived from the object's mean-motion derivative and \((da/dt)_{\mathrm{ref}}\) is measured from the reference propagation. State and STM propagation use a 60~s fourth-order Runge--Kutta step and are batched on GPU. Adaptive thresholds are evaluated at look-ahead times of 0, 1, 2, 3, 5, and 7 days using the propagated covariance history.

For object \(i\), the propagated covariance \(P_i(t)\) is rotated into the instantaneous RTN frame at each propagation step. The radial standard deviation is
\begin{equation}
\sigma_{R,i}(t)=
\sqrt{\left[P_i^{\mathrm{RTN}}(t)\right]_{11}}
\label{eq:radial_sigma}
\end{equation}
where \(\left[P_i^{\mathrm{RTN}}(t)\right]_{11}\) is the upper-left entry of the RTN covariance matrix and represents the radial position variance.

Because the instantaneous radial uncertainty varies with orbital phase, we take the local maximum of \(\sigma_{R,i}(t)\) over a \(\pm0.05\)-day (\(\pm72\)-minute) window centered at each look-ahead time. The resulting 144-minute window spans approximately one LEO orbital period or more, reducing sensitivity to the orbital phase at which the covariance is sampled. We denote this radial uncertainty envelope by \(\sigma_{R,i}^{\mathrm{env}}(t)\).

We define the adaptive apsis threshold \(T_i(t)\) to be
\begin{equation}
T_i(t)=k\,\sigma_{R,i}^{\mathrm{env}}(t),
\qquad k=2
\label{eq:adaptive_threshold}
\end{equation}
and set \(k=2\), corresponding to a two-standard-deviation scaling of the propagated radial uncertainty. Unlike the fixed \SI{50}{\kilo\meter} baseline, the adaptive threshold widens for objects with greater propagated uncertainty and narrows for well-characterized objects, allowing confidently separated pairs to be excluded from further analysis.

Some catalog objects lack sufficient ephemeris history to construct \(P_{i,0}^{\mathrm{RTN}}\). These objects receive a fixed \SI{50}{\kilo\meter} threshold~\cite{stevenson_2023}. This fallback ensures a defined screening threshold for every catalog object while preserving adaptive thresholding for objects with sufficient public orbital history.

\subsection{Apsis Envelope and Binary-Search Altitude Sweep}

To implement the apsis sieve, nominal object trajectories are propagated with SGP4 over the \SI{7}{\day} screening window with a \SI{30}{\minute} sampling interval. At each sample, the propagated Cartesian state is converted to osculating Keplerian elements, from which perigee and apogee altitudes are computed. For object \(i\), the minimum perigee \(q_i\) and maximum apogee \(Q_i\) over the propagated window define its nominal, uncertainty-free radial envelope. We expand this envelope by a fixed \SI{5}{\kilo\meter} on both sides as numerical padding. This padding is applied identically to the fixed and adaptive screening methods. The same padded nominal envelope is used at all evaluated look-ahead times, while the adaptive threshold \(T_i(t)\) varies with propagated radial uncertainty.

For a given look-ahead time \(t\), the expanded interval bounds for object \(i\) are
\begin{equation}
\begin{aligned}
\ell_i(t) &= q_i-\delta_{\mathrm{apsis}}-T_i(t),\\
h_i(t) &= Q_i+\delta_{\mathrm{apsis}}+T_i(t).
\end{aligned}
\label{eq:expanded_bounds}
\end{equation}
where \(\delta_{\mathrm{apsis}}=\SI{5}{\kilo\meter}\) is the fixed numerical
padding applied to the nominal apsis envelope, consistent with the
\SI{5}{\kilo\meter} conjunction-detection threshold used in prior
apogee--perigee filtering studies~\cite{woodburn_2009}. A pair of objects survives the apsis sieve when their expanded intervals overlap. A direct implementation evaluates the overlap condition for all \(N(N-1)/2\) object pairs. Algorithm~\ref{alg:apsis_sweep} instead sorts the expanded intervals by their lower bounds and uses binary search to reduce the interval-search complexity from \(O(N^2)\) pairwise comparisons to \(O(N\log N)\).

\begin{algorithm}
\caption{Binary-Search Altitude Sweep}
\label{alg:apsis_sweep}
\begin{algorithmic}[1]
\Require Expanded intervals
\(\{[\ell_i(t),h_i(t)]\}_{i=1}^{N}\)
\Ensure Candidate pair set \(\mathcal{C}\)

\State Sort intervals by increasing \(\ell_i(t)\), retaining object identifiers
\State \(\mathcal{C} \gets \emptyset\)

\For{\(i=1,\ldots,N-1\)}
    \State \(j_{\max} \gets
    \Call{UpperBound}{\{\ell_j(t)\},h_i(t)} - 1\)
    \For{\(j=i+1,\ldots,j_{\max}\)}
        \State \(\mathcal{C} \gets
        \mathcal{C}\cup\{(\mathrm{id}_i,\mathrm{id}_j)\}\)
    \EndFor
\EndFor

\State \Return \(\mathcal{C}\)
\end{algorithmic}
\end{algorithm}

Here, \Call{UpperBound}{} returns the first sorted index whose lower bound exceeds \(h_i(t)\). Because the intervals are sorted by their lower bounds, once
\(\ell_j(t)>h_i(t)\), no later interval can overlap interval \(i\).
For every interval considered before this cutoff, the complementary
overlap condition is satisfied automatically. Algorithm~\ref{alg:apsis_sweep}
therefore returns the same candidate set as exhaustive evaluation of
the apsis-overlap condition in Eq.~\ref{eq:apsis_overlap}. 
\section{Experimental Design}
\label{sec:experimental_design}

We evaluate the proposed adaptive apsis sieve along three primary dimensions. First, we measure how much it reduces the candidate-pair workload relative to brute-force screening and a fixed \SI{50}{\kilo\meter} apsis threshold. Second, we evaluate whether the adaptive thresholds preserve independently identified close approaches. Third, we characterize the computational scaling of the binary-search altitude sweep as catalog size increases. Supporting validation of the synthetic covariance estimates and the selected ephemeris-history length is provided in Appendix~\ref{app:cara_validation}.

All experiments were run on a workstation equipped with a 64-core AMD EPYC 9554 processor, 512~GB of RAM, and two NVIDIA L40 GPUs with 48~GB of memory each. Full-catalog covariance and state-transition-matrix propagation used a single L40 GPU with fixed batches of 1,500 objects and \texttt{float32} arithmetic, while runtime-scaling experiments were run CPU-only and single-threaded on the same machine.

\subsection{Catalog Data and Reference Configuration}

Catalog experiments use a Space-Track GP snapshot from 2026-05-30T16:17:02Z together with up to eight preceding ephemeris records per object, selected from the preceding 28 days of GP history~\cite{SpaceTrack}. Records are deduplicated by NORAD identifier and ephemeris epoch, yielding a propagated apsis catalog of 33,791 valid objects. Usable seven-day synthetic covariances are generated for 25,284 objects; the remaining 8,507 objects use the fixed \SI{50}{\kilo\meter} fallback threshold due to insufficient ephemeris history. We sample the propagated covariance at look-ahead times of 0, 1, 2, 3, 5, and 7 days and use the corresponding radial uncertainty to compute \(T_i(t)\) with \(k=2\).

For covariance propagation, we use GPU-batched computations implemented with astrojax. To verify the correctness of this implementation, we cross-check the propagated state-transition matrices and covariance evolution for selected objects against an independent implementation using the open-source Brahe astrodynamics library~\cite{brahe}. The GPU-based astrojax implementation is used for all full-catalog covariance propagation and associated timing results.

\subsection{Primary Screening and Recall Experiments}
\label{sec:screening_experiments}

The primary experiments evaluate two properties of the adaptive apsis sieve: its ability to reduce the candidate-pair workload and its ability to preserve physically relevant close approaches. Close-approach recall is evaluated using independently propagated reference sets containing pairs of satellites whose minimum sampled three-dimensional separation over the seven-day screening horizon is below \SI{5}{\kilo\meter}. We measure the fraction of these reference pairs retained by both the fixed and adaptive apsis sieves.

The full-catalog screening experiment compares three candidate populations: all \(N(N-1)/2\) unordered object pairs, the pairs retained by the fixed \SI{50}{\kilo\meter} apsis sieve, and the pairs retained by the proposed adaptive covariance-threshold sieve. We report the total number of retained candidate pairs, retained fraction, reduction in candidate-pair count relative to brute-force all-pairs screening, and additional reduction relative to the fixed-threshold sieve. One-vs-catalog performance is summarized using the median and 95th-percentile candidate reduction and query latency.

Algorithmic correctness of the binary-search implementation is evaluated separately by comparing its candidate set with exhaustive quadratic evaluation of the same apsis intervals and requiring exact agreement.

To illustrate how the aggregate workload reduction arises from object-specific thresholds, we examine three representative cases: a low-uncertainty payload, a rocket body, and a debris object whose seven-day radial threshold exceeds \SI{50}{\kilo\meter}. These examples illustrate the two-sided behavior of the method: thresholds can become substantially smaller than the fixed baseline when propagated uncertainty is low and can widen beyond \SI{50}{\kilo\meter} when greater uncertainty warrants more conservative screening.

\subsection{Runtime Scaling}
\label{sec:runtime-scaling}

To isolate the computational scaling of the altitude-sweep algorithm from orbit propagation and covariance generation, we benchmark the screening step separately using precomputed apsis intervals. We evaluate seven catalog sizes from \(N=500\) to \(N=33{,}777\), using five random subsets at each size and five timed repetitions after one warm-up run. All timing experiments are performed CPU-only and single-threaded on the same hardware.

\section{Results}
\label{sec:results}

\subsection{Screening Recall and Candidate Reduction}

The 33,791-object catalog contains at most 570,898,945 possible conjuncting pairs. \Cref{tab:screening-results} gives the primary screening result: the adaptive sieve retains 162,819,449 pairs at seven days, 28.78\% fewer than the fixed 50~km sieve and 71.48\% fewer than brute force.

We evaluate close-approach retention using the reference sets described in \Cref{sec:screening_experiments}. The SGP4-propagated and numerical-propagation sets contain 98 and 1,768 pairs, respectively, with sampled three-dimensional separation below \SI{5}{\kilo\meter} over the seven-day horizon. Both the fixed \SI{50}{\kilo\meter} sieve and the adaptive sieve retain every reference pair in both sets, giving 100\% recall. Together, these results show that the adaptive sieve reduces the candidate set substantially while preserving all close approaches in both reference sets.

The binary-search altitude sweep also returns exactly the same candidate set as exhaustive evaluation of the same expanded apsis intervals. The sweep therefore introduces no additional pair loss beyond that imposed by the screening thresholds themselves.
For one-vs-catalog screening, the median candidate reduction is 26.00\%, with a 95th-percentile reduction of 59.63\%. Because the adaptive thresholds are not capped at \SI{50}{\kilo\meter}, 4.87\% of objects return more candidates than under fixed screening. Median conjunction-screening query latency is \SI{50.46}{\micro\second}, with a 95th-percentile latency of \SI{81.92}{\micro\second} over 1,000 sampled queries.

\begin{table*}[t]
\centering
\caption{Seven-day all-vs-all screening results. Reduction is reported relative to the unscreened pair universe.\label{tab:screening-results}}
\begin{tabular}{@{}lrrr@{}}
\toprule
Method & Pairs retained & Retained fraction & Reduction vs. brute force\\
\midrule
Brute force & 570,898,945 & 100.00\% & 0.00\%\\
Fixed 50 km & 228,619,849 & 40.05\% & 59.95\%\\
Adaptive & 162,819,449 & 28.52\% & 71.48\%\\
\bottomrule
\end{tabular}
\end{table*}

\begin{figure*}[t]
\centering
\makebox[\textwidth][c]{%
\subfloat[Centispace payload]{%
    \includegraphics[width=0.30\textwidth]
    {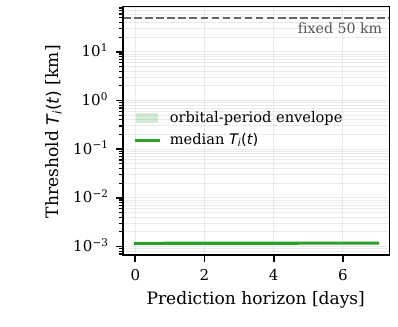}%
    \label{fig:threshold_payload}%
}%
\hfill
\subfloat[SL-14 rocket body]{%
    \includegraphics[width=0.30\textwidth]
    {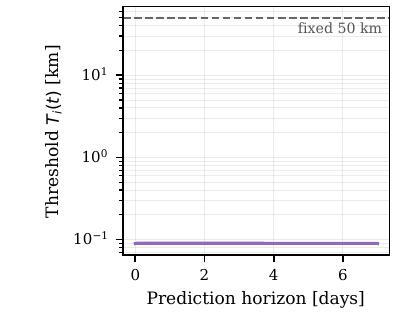}%
    \label{fig:threshold_rocket_body}%
}%
\hfill
\subfloat[Fregat debris]{%
    \includegraphics[width=0.30\textwidth]
    {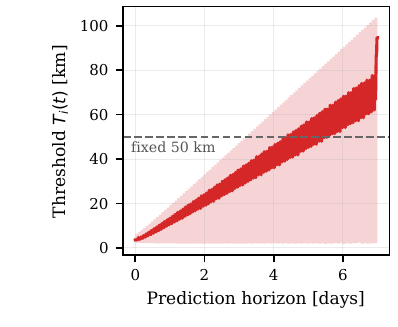}%
    \label{fig:threshold_debris}%
}%
}

\caption{Adaptive-threshold evolution for three representative objects. Solid curves show the orbital-window median and shading shows the local min--max envelope. The dashed line marks the fixed \SI{50}{\kilo\meter} baseline. Logarithmic vertical axes are used for the payload and rocket-body cases, while a linear vertical axis is used for the debris case.}
\label{fig:threshold_examples}
\end{figure*}
\subsection{Object-Specific Screening Conservatism}

The reduction in candidate pairs arises from object-specific differences in the adaptive threshold. \Cref{fig:threshold_examples} shows the threshold evolution for three representative objects. The Centispace payload and SL-14 rocket body remain well below the fixed \SI{50}{\kilo\meter} baseline throughout the seven-day horizon (\Cref{fig:threshold_payload,fig:threshold_rocket_body}), reflecting the small propagated radial uncertainty for these objects. Their narrower altitude intervals allow additional non-overlapping pairs to be rejected relative to fixed screening.

The Fregat debris object exhibits the opposite behavior (\Cref{fig:threshold_debris}). Its propagated radial uncertainty grows substantially over the prediction horizon, causing the adaptive threshold to exceed the fixed \SI{50}{\kilo\meter} baseline. The adaptive sieve therefore preserves a wider altitude interval for this object rather than applying the tighter threshold used for the lower-uncertainty cases. Together, these examples illustrate the intended object-specific behavior of the method: tighter screening when propagated uncertainty is small and more conservative screening when it is large. These examples characterize individual threshold behavior rather than the catalog-wide threshold distribution.

\subsection{Screening Scaling}

\begin{figure*}[t]
\centering
\makebox[\textwidth][c]{%
\subfloat[Candidate-pair scaling]{%
    \includegraphics[width=0.47\textwidth]
    {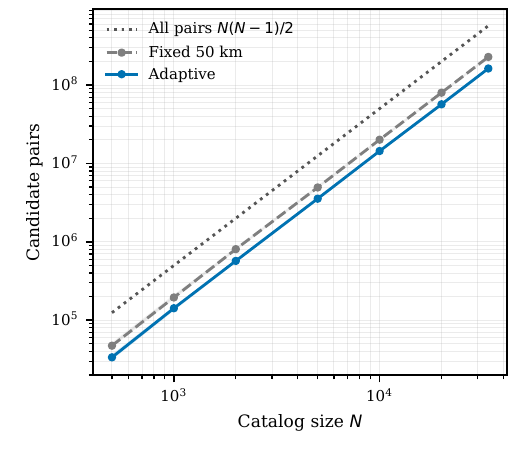}%
    \label{fig:scaling_candidates}%
}%
\hfill
\subfloat[CPU screening-time scaling]{%
    \includegraphics[width=0.47\textwidth]
    {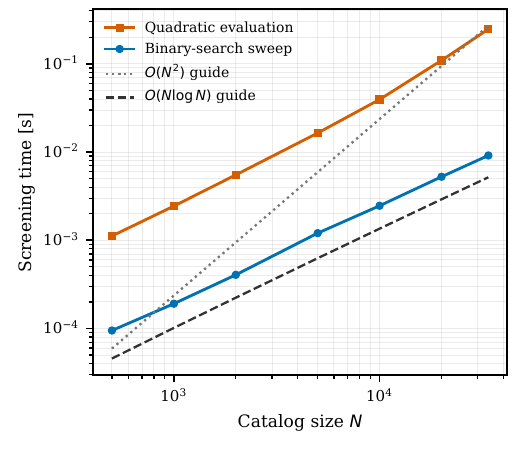}%
    \label{fig:scaling_runtime}%
}%
}

\caption{Scaling of candidate-pair count and CPU screening time with catalog size. The \(O(N^2)\) and \(O(N\log N)\) curves are theoretical scaling guides rather than fitted models. Orbit propagation and covariance generation are excluded.}
\label{fig:scaling}
\end{figure*}

The reduction in candidate pairs persists as the catalog size increases. As shown in \Cref{fig:scaling_candidates}, the adaptive sieve retains fewer candidate pairs than the fixed \SI{50}{\kilo\meter} sieve at every tested catalog size, while both remain below the unscreened \(N(N-1)/2\) pair population. This indicates that the altitude sieve maintains its reduction in candidate pairs for a given catalog even with growth in the satellite population.

The runtime advantage of the altitude sweep also increases with catalog size. Under the CPU-only benchmark described in \Cref{sec:runtime-scaling}, screening \(N=33{,}777\) objects requires \SI{0.00916}{\second} with the sweep, compared with \SI{0.247}{\second} for exhaustive quadratic interval evaluation, a \(27\times\) speedup. The theoretical \(O(N^2)\) and \(O(N\log N)\) curves in \Cref{fig:scaling_runtime} are shown only as scaling references; the measured timings demonstrate the increasing separation between exhaustive pairwise evaluation and the indexed sweep over the tested catalog sizes.

Appendix~\ref{app:covariance_validation} provides supporting analyses of the synthetic covariance estimator, including sensitivity to ephemeris-history length and comparison with NASA CARA covariance data.

\section{Discussion}
The main benefit of the adaptive sieve comes from replacing a catalog-wide assumption about uncertainty with an object-specific one. A fixed apsis threshold treats every orbit estimate as equally uncertain, even though the propagated radial uncertainty can differ substantially across objects. The representative cases in \Cref{fig:threshold_examples} illustrate both directions of this effect: well-characterized objects can be screened with substantially narrower intervals, while an uncertain debris object can receive a threshold wider than the fixed \SI{50}{\kilo\meter} baseline. The candidate-pair reduction arises from assigning the screening interval according to the uncertainty inferred for each object.

The recall experiments provide a separate check on whether the additional filtering removes physically relevant encounters. Both independently propagated reference sets retain all identified close approaches under the adaptive sieve. This result supports using the adaptive threshold as a first-stage screening criterion: it reduces the number of pairs passed downstream without losing any of the close approaches observed in the evaluated reference sets.

The binary-search altitude sweep keeps this additional flexibility computationally tractable. For a fixed set of expanded apsis intervals, the sweep returns the same candidate set as exhaustive pairwise interval evaluation while avoiding comparisons with intervals that cannot overlap. The measured runtime gap grows with catalog size, reaching a \(27.0\times\) speedup at the largest tested catalog. This means the object-specific thresholds can be evaluated without reintroducing the full cost of quadratic interval testing.

The present results support the use of public-ephemeris disagreement to size radial screening thresholds, but they do not establish a calibrated probability model for orbit uncertainty. The synthetic covariance is built from related public solutions propagated with the same underlying model, so shared errors and temporal correlation remain in the residuals. The chosen values of \(k=2\) and \(n_{\mathrm{recent}}=8\) should therefore be viewed as screening parameters for this implementation rather than universal uncertainty calibrations. The evaluation is also limited to LEO and a seven-day horizon, which defines the regime in which the reported screening behavior has been demonstrated.
\section{Conclusions}
We introduced an uncertainty-aware apsis screening method that replaces a fixed catalog-wide threshold with object-specific thresholds derived from propagated synthetic covariance. Public ephemeris disagreement is converted into an object-specific covariance estimate, propagated over the screening horizon, and reduced to a radial threshold that can tighten or widen according to each object's predicted uncertainty. To evaluate these heterogeneous screening intervals computationally efficiently at catalog scale, we use a binary-search altitude sweep that preserves the same candidate set as exhaustive evaluation. 

On a 33,791-object Space-Track catalog, the adaptive sieve retains 162.8 million candidate pairs over a seven-day horizon, reducing the candidate set by 28.78\% relative to the fixed \SI{50}{\kilo\meter} baseline and by 71.48\% relative to brute-force all-pairs screening. Both independently propagated reference sets retain 100\% of identified close approaches. The binary-search sweep returns the same candidate set as exhaustive interval evaluation and achieves a 27 times speedup at the largest tested catalog size.


These results suggest that public ephemeris history can support more selective first-stage conjunction screening without applying the same conservatism to every catalog object. As tracked populations continue to grow, reducing the number of pairs passed to higher-fidelity propagation and collision-risk analysis can lower the computational burden of catalog-scale screening while preserving conservative treatment of uncertain objects. Future work should calibrate the threshold scale against additional independent close-approach data, evaluate performance across multiple catalog epochs and orbital regimes, and determine whether the history length and threshold scaling should vary by object.

\acknowledgements

This research was supported in part by the NASA Ames Center Innovation Fund, the Fannie and John Hertz Foundation, and the National Defense Science and Engineering Graduate (NDSEG) Fellowship Program. Specifically, this material is based upon work supported by the Air Force Office of Scientific Research under award number FA9550-25-C-B010 in the amount of currently negotiated tuition and stipend rates. 

\bibliographystyle{IEEEtranN}
\bibliography{refs}
\thebiography

\begingroup
\setkeys{Gin}{keepaspectratio}

\begin{biographywithpic}{Vedant Srinivas}{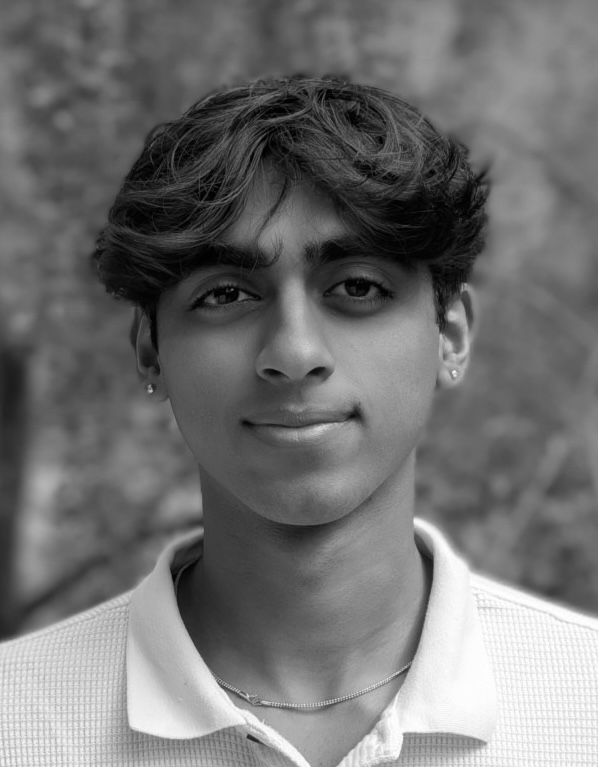}
is an undergraduate in the Stanford Intelligent Systems Laboratory in the department of Computer Science at Stanford University. His research interests include optimization and decision-making for autonomous systems, with experience as an AI Software Engineer Intern at Salesforce AI Research. He is also the co-founder of IyarkAI, a startup deploying real-time wildlife monitoring systems with the Washington Department of Transportation, and previously worked at the UC Davis Road Ecology Center on edge-AI systems for wildlife detection and collision prevention.
\end{biographywithpic}

\begin{biographywithpic}{Grace Ra Kim}{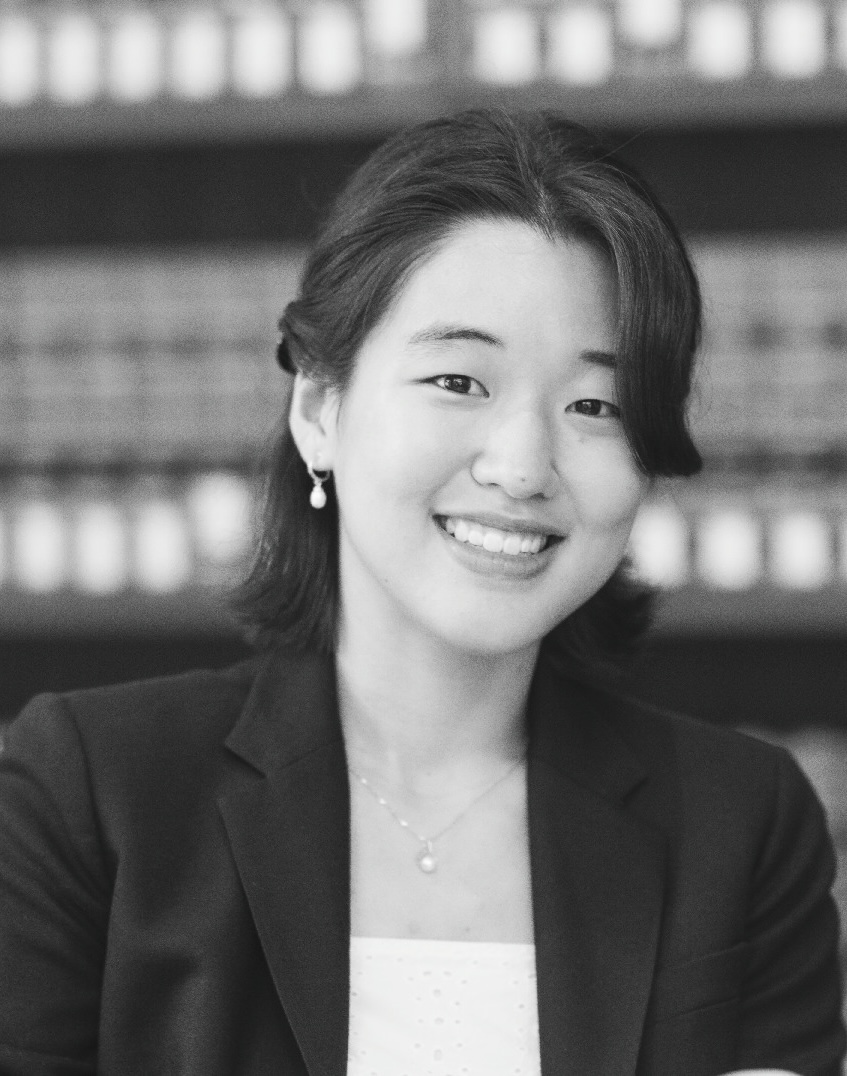}
is a PhD candidate in the Stanford Intelligent Systems Laboratory (SISL) in the Department of Aeronautics and Astronautics at Stanford University, as a Hertz and NDSEG Fellow. She received her M.Phil. in Advanced Computer Science at the University of Cambridge as a Marshall Scholar in 2024, and graduated cum laude with high honors in a B.S. in Engineering Sciences from Harvard University in 2023. Her research interests lie at the intersection of space traffic management, safe autonomous systems in space, and space policy.
\end{biographywithpic} 

\begin{biographywithpic}{Duncan Eddy}{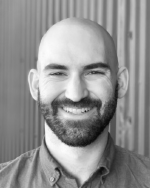}
is a staff research fellow in the Stanford Intelligent Systems Laboratory (SISL) and the Executive Director of the Center for AI Safety at Stanford University. His research focuses on decision-making in safety-critical, climate, and space systems. He received a B.S. in Mechanical Engineering from Rice University in 2013 and a Ph.D. in Aerospace Engineering from Stanford University in 2021. Prior to returning to Stanford, he was the Director of Space Operations at Capella Space Corporation, where he built a fully automated constellation tasking and delivery system. He later founded and led the Constellation Management and Space Safety Organization at Amazon's Project Kuiper.
\end{biographywithpic}

\begin{biographywithpic}{Mykel J. Kochenderfer}{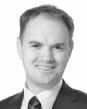}
is an Associate Professor of Aeronautics and Astronautics and Associate Professor, by courtesy, of Computer Science at Stanford University. He is the director of the Stanford Intelligent Systems Laboratory (SISL), conducting research on advanced algorithms and analytical methods for the design of robust decision-making systems. Prior to joining the faculty in 2013, he was at MIT Lincoln Laboratory, where he worked on airspace modeling and aircraft collision avoidance. He received his Ph.D. from the University of Edinburgh in 2006, where he studied at the Institute of Perception, Action and Behaviour in the School of Informatics. He received B.S. and M.S. degrees in computer science from Stanford University in 2003.
\end{biographywithpic}
\endgroup

\appendices
\section{Synthetic Covariance Validation}
\label{app:covariance_validation}

This appendix provides supporting validation for the synthetic covariance model used to construct the adaptive apsis thresholds. We first compare the propagated synthetic covariance with independently reported NASA CARA covariance, then evaluate sensitivity to the number of preceding ephemeris records used in the estimator, and finally examine representative RTN uncertainty propagation over the seven-day screening horizon.

\subsection{Comparison with NASA CARA Covariance}
\label{app:cara_validation}

We compare the propagated synthetic covariance against the NASA CARA PcTestCaseCDMs dataset, which provides CCSDS conjunction data messages with reported radial--transverse--normal (RTN) covariance at time of closest approach (TCA)~\cite{nasa_cara_tools}. For each object-case, only GP element sets issued before TCA are used. The initial synthetic covariance is constructed from the available historical solutions and propagated to TCA using the procedure described in Section~\ref{sec:methodology}. Cases outside the LEO regime or without sufficient public ephemeris history are excluded.

Because the adaptive apsis threshold depends on radial uncertainty, we focus the primary comparison on the radial variance. Let \(\widehat{P}_{RR}\) denote the propagated synthetic radial variance at TCA and \(P_{RR}^{\mathrm{CARA}}\) denote the corresponding reported CARA value. We define the symmetric agreement factor as
\begin{equation}
f_R=
\max\left(
\frac{\widehat{P}_{RR}}{P_{RR}^{\mathrm{CARA}}},
\frac{P_{RR}^{\mathrm{CARA}}}{\widehat{P}_{RR}}
\right),
\label{eq:cara_radial_factor}
\end{equation}
where \(f_R=1\) indicates exact agreement.

\begin{figure*}[t]
\centering
\makebox[\textwidth][c]{%
\subfloat[Case-level comparison]{%
    \includegraphics[width=0.46\textwidth]{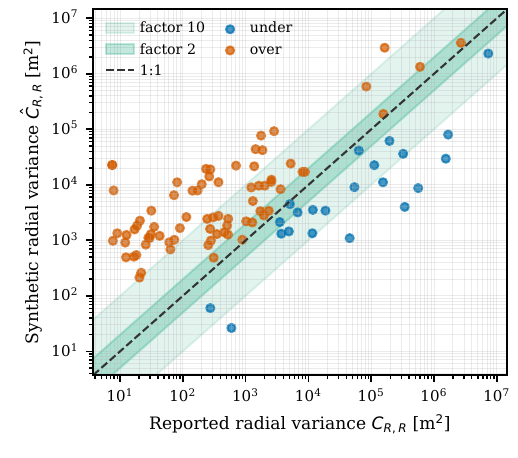}%
    \label{fig:cdm_cases}%
}%
\hfill
\subfloat[Agreement distribution]{%
    \includegraphics[width=0.46\textwidth]{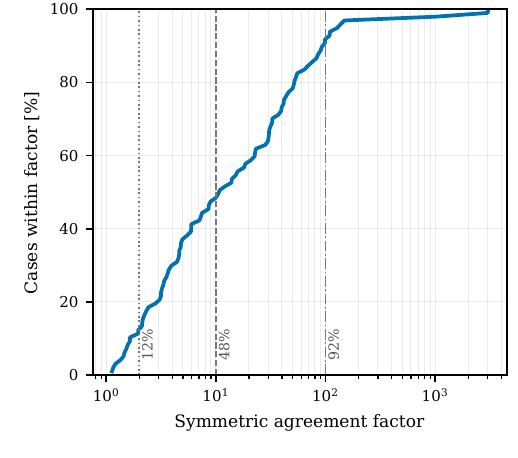}%
    \label{fig:cdm_agreement}%
}%
}
\caption{Comparison of propagated synthetic and reported radial variance at TCA for 97 LEO object-cases from the NASA CARA \textsc{PcTestCaseCDMs} dataset. The case-level comparison shows exact agreement and factor-of-two and factor-of-ten regions, while the agreement distribution summarizes the symmetric factor between the synthetic and reported radial variances.}
\label{fig:cdm}
\end{figure*}

The CARA corpus yields 98 usable LEO object-cases, of which 97 have finite positive radial-variance ratios. \Cref{fig:cdm_cases} compares the propagated synthetic and reported radial variances at TCA, while \Cref{fig:cdm_agreement} summarizes the corresponding agreement factors. The median synthetic-to-reported radial variance ratio is 5.95. In the 97 finite cases, 12.4\% lie within a factor of two, 48.5\% lie within a factor of ten, and 91.8\% lie within a factor of one hundred of the reported CARA radial variance. The broad spread indicates that the synthetic covariance provides an object-specific screening scale, but does not reproduce operator-grade covariance uniformly across individual cases.

The axis-specific results in \Cref{tab:covariance-agreement} show that the agreement differs substantially across the RTN components. Along-track variance is underpredicted in 70.1\% of cases, whereas the radial and cross-track components are overpredicted on median. This directional variation motivates using the radial component directly for the apsis threshold rather than treating the synthetic covariance as a uniformly calibrated full-state covariance estimate.

\begin{table*}[t]
\centering
\caption{Agreement between propagated synthetic and reported CARA covariance diagonal entries.}
\label{tab:covariance-agreement}
\begin{tabular}{lrrrrrr}
\toprule
Axis & Cases & Median ratio & Underpredicted & Within $2\times$ & Within $10\times$ & Within $100\times$ \\
\midrule
Radial      & 97 & 5.95  & 22.7\% & 12.4\% & 48.5\% & 91.8\% \\
Along-track & 97 & 0.18  & 70.1\% & 16.5\% & 49.5\% & 66.0\% \\
Cross-track & 97 & 21.77 & 11.3\% & 17.5\% & 38.1\% & 76.3\% \\
\bottomrule
\end{tabular}
\end{table*}

\subsection{Ephemeris-History Sensitivity}
\label{app:history_sensitivity}

\begin{figure*}[t]
\centering
\makebox[\textwidth][c]{%
\subfloat[Well-tracked payload]{%
    \includegraphics[width=0.30\textwidth]{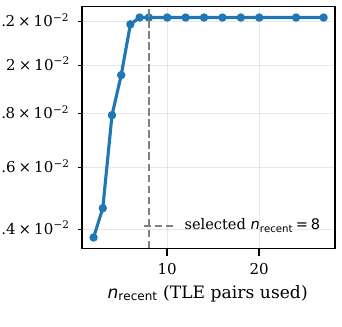}%
    \label{fig:history_payload}%
}%
\hfill
\subfloat[Median object]{%
    \includegraphics[width=0.30\textwidth]{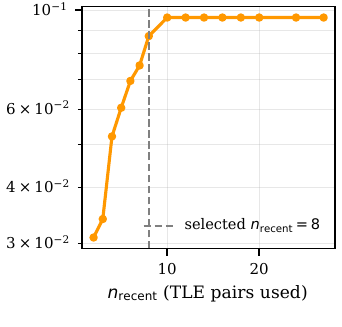}%
    \label{fig:history_median}%
}%
\hfill
\subfloat[Debris object]{%
    \includegraphics[width=0.30\textwidth]{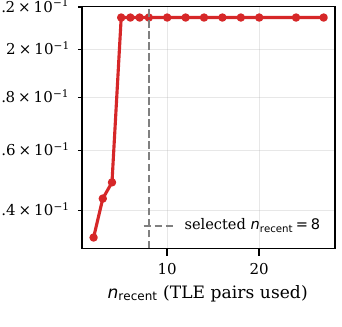}%
    \label{fig:history_debris}%
}%
}
\caption{Sensitivity of the seven-day adaptive threshold to the number of preceding ephemeris records used to construct the synthetic covariance. The dashed line marks the selected configuration, \(n_{\mathrm{recent}}=8\).}
\label{fig:history_ablation}
\end{figure*}

To evaluate the choice of \(n_{\mathrm{recent}}\), we recompute the synthetic covariance using progressively larger sets of preceding ephemeris records. \Cref{fig:history_ablation} shows the resulting seven-day adaptive threshold as \(n_{\mathrm{recent}}\) is increased for three representative objects. The thresholds approach stable values after approximately 5--10 preceding records, although the stabilization rate differs across the three cases.

The debris case in \Cref{fig:history_debris} reaches its plateau after approximately five preceding records. The well-tracked payload in \Cref{fig:history_payload} stabilizes near the selected value of \(n_{\mathrm{recent}}=8\), while the median-object case in \Cref{fig:history_median} continues to increase modestly beyond eight records before flattening. We therefore use \(n_{\mathrm{recent}}=8\) as a common configuration that captures most of the observed stabilization while avoiding a requirement for longer ephemeris histories.

\subsection{Representative RTN Uncertainty Propagation}
\label{app:rtn_propagation}

\begin{figure*}[t]
\centering
\makebox[\textwidth][c]{%
\subfloat[NORAD 12978, well-tracked payload]{%
    \includegraphics[width=0.31\textwidth]{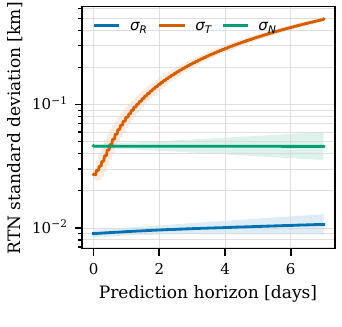}%
    \label{fig:covariance_payload}%
}%
\hfill
\subfloat[NORAD 17271, median object]{%
    \includegraphics[width=0.31\textwidth]{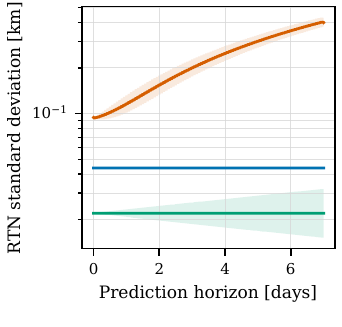}%
    \label{fig:covariance_median}%
}%
\hfill
\subfloat[NORAD 54487, debris]{%
    \includegraphics[width=0.31\textwidth]{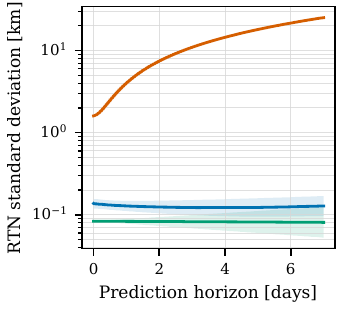}%
    \label{fig:covariance_debris}%
}%
}
\caption{STM-propagated RTN position uncertainty for three representative objects. Curves show the radial, along-track, and cross-track standard deviations over the seven-day prediction horizon, with shading indicating the corresponding orbital-period minimum--maximum envelope.}
\label{fig:covariance}
\end{figure*}

\Cref{fig:covariance} illustrates the STM-propagated RTN position standard deviations for three representative objects. Along-track uncertainty grows most strongly over the seven-day prediction horizon in all three examples, while the radial component used by the apsis sieve remains substantially smaller and varies more slowly. The magnitude of the propagated uncertainty also differs considerably across objects, illustrating the object-to-object variation that motivates an adaptive threshold.

\end{document}